\documentclass[12pt]{article}

\usepackage{verbatim}
\usepackage{enumerate}
\usepackage{color, float}
\usepackage{amssymb,amsthm,amsmath}
\usepackage{latexsym}
\usepackage{graphicx}
\usepackage{hyperref}
\usepackage{float}
\usepackage{algorithm}
\usepackage{algorithmic}
\usepackage{url}
\usepackage{float}
\usepackage{enumitem} 
\usepackage{subcaption}
\usepackage{url}
\usepackage{fullpage}
\usepackage{tikz}
\usepackage{tabularx}
\usetikzlibrary{arrows.meta, calc}
\usetikzlibrary{graphs, positioning, shapes.geometric}
\usetikzlibrary{decorations.markings}

\theoremstyle{definition}

\title{Topological Data Analysis and Graph-Theoretic Approaches for Tennis Match Prediction}

\author{
Jake Schwaderer, Alexander Bastien, Omid Khormali$^{}$\footnote{Corresponding author: ok16@evansville.edu} ,  Alejandro Navarrete, \\Mia Pesavento, Angelika Elderbrook\\
\small Department of Mathematics\\[-0.8ex]
\small University of Evansville\\[-0.8ex]
\small Evansville, Indiana 47722, USA.\\
	\small \texttt{js1104@evansville.edu, ab995@evansville.edu, ok16@evansville.edu,}\\ \small \texttt{an226@evansville.edu, mp367@evansville.edu, ae205@evansville.edu}
}

\date{}
\begin{document}

\maketitle

\begin{abstract}

We present two approaches for predicting tennis match outcomes using topological data analysis and graph theory on ATP singles matches from 2000–2025. The first method applies lower-star filtration to player competitive networks, extracting topological features through persistent homology using four summary methods (VAB, HNAV, HWNAV, OW-HNPV) combined with Modified Band Depth analysis. Algorithmic optimizations including ego graph approximations and triangle elimination enable analysis of about 66k matches. Our Random Forest model achieves 66.2\% accuracy (AUC = 0.719) using topological, graph-theoretic, and ranking features. Feature importance analysis reveals that rankings contribute 36.3\%, centralities 25.5\%, and TDA features 24.0\%, with topological features providing complementary signal. When rankings are unavailable, the topology-only model maintains 63.56\% accuracy, demonstrating that network-derived features alone capture meaningful competitive structure. The second method uses a modified Katz similarity index with temporal edge weighting, achieving 62.48\% accuracy on held-out test data. This work represents the first application of lower-star filtration to tennis prediction, provides systematic comparison of four topological summary methods in sports analytics, and demonstrates that TDA can achieve above-chance prediction using network topology alone while providing additional value when combined with traditional features.

\end{abstract}



\section{Introduction}
Sport data analysis is crucial for teams and competitive organizations because it provides valuable insight to enhance player and team performance. By examining large sets of performance data, teams will be able to identify strengths, weaknesses, and other useful patterns that may not be clear during a game or practice. Using sport data analysis, coaches can make real time-decisions, strategic planning, and create long-term team building and improvements. Beyond refining performance, sport data analysis also plays a key role in marketing and merchandise as it creates a more personalized and engaging experience for fans to follow along with. By comparing the data between teams and players, fans can pick a favorite support and route to feel like they are part of the game itself.\\

Machine learning is now at the forefront of advancing sport data analysis by being able to handle larger datasets, accurately predict outcomes, and compose better models for coaches and players to use. These models can now process variables such as previous matchup and player tendencies, long with real time data from the game. All of these added advantages result in more detailed predictions and reports. As machine learning and AI continues to evolve and improve, it becomes essential for understand complex patterns that were previously thought impossible to capture \cite{TAMU_KNSM}. \\

In this paper, we focus on tennis data analysis by applying topological data analysis, graph theory, and machine learning methods to predict match outcomes between two players based on their historical performance and competitive network structure.

\subsection{Summary of Study for Previous Tennis Data Analysis}

A 2005 study by Newton and Keller \cite{Newton2005} explored how probability theory can predict a player's chances of winning a game, set, or match. The study assumed that points are independent and identically distributed, and derived formulas based on a player's probability of winning a point on serve. Using data from the 2002 U.S. Open and Wimbledon, the predicted outcomes showed strong agreement with actual match results. The paper also proved that under their model, the probability of winning a set or match is independent of which player serves first.

In 2017 study, Jacob Gollub wrote “Producing Win Probabilities for Professional Tennis Matches from Any Score” \cite{Gollub2017} which explored how mathematical models can predict the chances of player winning a match at any period of play. The study used a hierarchical Markov model to combined pre-match expectations with real time time factors, such as current score and serve performance. This resulted in the system continuously updated the chance of each player winning. This approach demonstrated how statistical patterns in scoring and serve strength can be used to accurately predict outcomes before a match is finished.\\
 
 Another study in 2025 by Boyuan Li, Zihui Deng, and Gaurav Gupta, "Predicting tennis match outcomes mid‑game using machine learning on psychological and physical data"\cite{Li2025} explored how machine learning can predict the outcome of live tennis matches by analyzing psychological and physical data. The study used real-time factors such as player movements, body language and other psychological responses to record perforce trends and momentum shifts. By combining all of these factors, the model was able to predict the outcome of the live tennis match before it ended, demonstrating how ones physical condition and mental state and influence a competitive performance.\\

 \subsection{Topological Data Analysis and Persistence Diagram Summaries}
 
 Topological Data Analysis (TDA) is the analysis of how data is organized beyond simple pairwise relationships. The method was pioneered in the early 2000s through research by Herbert Edelsbrunner, David Letscher, and Afra Zomorodian \cite{EdelsbrunnerLetscherZomorodian2002} before becoming more popularized in 2009 through Gunnar Carlsson's paper ``Topology and Data'' \cite{Carlsson2009}. \\
 
 Persistent homology, the core tool in TDA, tracks topological features across multiple scales of data by computing \textit{Betti numbers}, topological invariants that count connected components ($\beta_0$), loops or cycles ($\beta_1$), and voids ($\beta_2$) at each filtration level \cite{Edelsbrunner2010}. These features are encoded in persistence diagrams, which represent the birth and death times of topological features as the filtration parameter varies. However, persistence diagrams are not directly compatible with standard machine learning algorithms, motivating the development of vectorized summaries.\\
 
 Several methods have been proposed to convert persistence diagrams into fixed-dimensional vectors suitable for statistical analysis. Persistence landscapes \cite{Bubenik2015} represent diagrams as sequences of piecewise-linear functions, while persistence images \cite{Adams2017} create smoothed 2D representations in birth-persistence coordinates. For graph-structured data, the Vector of Averaged Bettis (VAB) \cite{Islambekov2024} provides a particularly efficient approach by discretizing and averaging the Betti function over scale intervals, yielding interpretable vectors that capture the cumulative presence of topological features.\\
 
 Recently, Khormali \cite{Khormali2024} introduced the Overlap-Weighted Hierarchical Normalized Persistence Velocity (OW-HNPV), a velocity-based alternative that measures the \textit{rate} at which topological features appear and disappear rather than their cumulative presence. Unlike VAB and other static summaries, OW-HNPV weights features by their actual overlap with hierarchical scale intervals, automatically downweighting short-lived noise while capturing temporal dynamics. This velocity-based perspective has proven particularly effective for anomaly detection in time-varying networks \cite{Khormali2024}, making it well-suited for analyzing the dynamic structure of competitive sports networks.\\
 
 Our paper represents one of the first applications of TDA to tennis match prediction, combining multiple topological summary methods (VAB, HNAV, HWNAV, OW-HNPV) with Modified Band Depth (MBD) analysis to extract predictive features from player competitive networks.

\section{Method}\label{methods}
We present two complementary approaches for tennis match prediction: a lower-star filtration method based on topological data analysis, and a graph-theoretic approach using a modified Katz similarity index.\\

The data we are using is sourced from a user on the website Kaggle.com that includes all the match data from singles ATP level tournaments since the year 2000. It includes the names of the players, who won the set, the tournament date and name, the score of the sets, among other things. The dataset we have was last updated right after the 2025 US Open, however more updated data is available, and the dataset is under the public domain.

Player rankings were obtained from the official ATP ranking system contemporaneous with each match. ATP rankings are computed weekly using a rolling 52-week points system based on tournament performance. The ranking difference feature $\Delta\text{rank} = \text{rank}(p_1) - \text{rank}(p_2)$ uses rankings as of the match date, ensuring no temporal data leakage in historical analysis. ATP rankings use an inverse scale where lower numerical values indicate stronger players (rank 1 = world's best player). For deployment on future matches, current rankings must be obtained from the ATP official rankings database and updated weekly.

\subsection{Lower-Star Filtration}
    Lower-star filtration is a fundamental method in topological data analysis for constructing nested sequences of simplicial complexes from graph-structured data \cite{Edelsbrunner2010}. Given a graph $G = (V, E)$ with a real-valued function $g: V \to \mathbb{R}$ defined on its vertices, the lower-star filtration builds a sequence of sublevel sets by progressively including simplices (vertices, edges, triangles, etc.) based on the maximum function value among their constituent vertices \cite{Edelsbrunner2010, Carlsson2009}.\\
    
    Formally, we first enhance the graph $G$ to include higher-dimensional simplices formed by cliques in the graph (e.g., triangles from 3-cliques, tetrahedra from 4-cliques). The function $g$ is extended to all simplices by defining $g(\sigma) = \max_{v \in \sigma} g(v)$, where $\sigma$ represents a simplex and $v$ ranges over its vertices. The sublevel set at filtration value $t$ is then defined as:
    \begin{equation}
    	K_t = \{\sigma : g(\sigma) \leq t\}.
    \end{equation}
    
    As $t$ increases from $\min_{v \in V} g(v)$ to $\max_{v \in V} g(v)$, this creates a nested sequence of simplicial complexes $K_{t_1} \subseteq K_{t_2} \subseteq \cdots \subseteq K_{t_m}$, called a filtration. Persistent homology \cite{Edelsbrunner2008, EdelsbrunnerLetscherZomorodian2002, Zomorodian2005} tracks topological features (connected components, loops, voids) as they appear and disappear across this filtration, encoding this information in a persistence diagram that summarizes the multiscale topological structure of the data.\\
    
    In our application to tennis match prediction, the function $g(v)$ represents a temporally-weighted measure of player strength (defined in Section~\ref{gfunction}), and the resulting persistence diagrams capture the hierarchical competitive structure within induced subgraphs of player matchups. The lower-star filtration is particularly well-suited for this application because it naturally incorporates the vertex-level information (player strength) while revealing higher-order topological patterns in the competitive network \cite{Islambekov2024}.
    
\subsubsection{The g(v) Function} \label{gfunction}
    Before we can compute lower-star filtration, we need to assign a value to every node, edge, and triangle in the graph. For the edges and triangles, we simply take the largest value from the nodes that comprise them. \\
    For the g function value of a node, we decided to base it off the overall win rate of a player. However, that caused many issues, namely players from the oldest years in the dataset could be predicted above those from the most recent year. As we are trying to predict a match as if it will happen right now, we decided to add a weight to the g function based off what we deemed as the most successful year for a given player. However, that still left the issue of players from the most recent year who won two or three sets, and only lost once as they played in in tournament. To balance that out we added in that players shares of overall wins so those who only played a few sets would not be overestimated. With that we can finally calculate the g function. \\
    
    Given that $v_p$ is the total number of wins from a player, $m_p$ is the total number of matches played by a player, and $m$ is the total number of matches in the dataset; We can define the g function as:
    \[g(v) = \frac{v_p^2}{m_pm}w\]
    $w$, or the weight, is given by the following equations given that $0 \leq Y \leq 25$, $v_{pY}$ is the number of wins in a given year for a player, and that $m_{pY}$ is the number of matches played by a player in a given year:
    \[w = \frac{1+e^{-3.25}}{1+e^{0.5*(18.5-y)}}, y= \text{Y of }max(\frac{v_{pY}m_{pY}}{v_pm_p})\]
    Finally, we divide every g function value by the largest value so we can have every value be within the interval $g(v) = [0, 1]$ to complete the filtration and subsequent analysis.

\subsubsection{Topological Feature Extraction and Modified Band Depth}
    
    For each match between players $p_1$ and $p_2$, we construct an induced subgraph $H$ by extracting the ego graphs (local neighborhoods) around both players and taking their union. Starting with radius $r = 2$, we incrementally increase $r$ until the induced subgraph contains at least 75 nodes, ensuring sufficient topological structure for analysis. We then compute lower-star filtration on $H$ using the $g(v)$ values, yielding persistence diagrams for dimensions $k \in \{0, 1\}$.\\
    
    To convert these persistence diagrams into predictive features, we compute four different topological summaries: Vector of Averaged Bettis (VAB) \cite{Islambekov2024}, Hierarchical Normalized Averaged Velocity (HNAV), Hierarchical Weighted Normalized Averaged Velocity (HWNAV), and Overlap-Weighted Hierarchical Normalized Persistence Velocity (OW-HNPV) \cite{Khormali2024}. Each method produces a fixed-dimensional vector representation of the persistence diagram, discretized into 999 time steps with $\Delta t = 0.001$.\\
    
    We additionally compute Modified Band Depth (MBD) scores to quantify how typical each match's topological signature is relative to a reference distribution. For each match, we construct nine reference graphs: six ego graphs centered on each player at radii $r \in \{2, 3, 4\}$, and three induced subgraphs from randomly selected player pairs (held constant across all predictions). The MBD score for the target graph $X_m$ is computed as:
    \begin{equation}
    	\text{MBD}(X_m) = \frac{1}{\binom{9}{2}} \sum_{1 \leq i < j \leq 9} \frac{1}{999} \sum_{s=1}^{999} {1}_{B_{ij}(s)}(X_m(s)),
    \end{equation}
    where $B_{ij}(s) = [\min(X_i(s), X_j(s)), \max(X_i(s), X_j(s))]$ defines the band at position $s$, and ${1}_{B_{ij}(s)}(X_m(s))$ is the indicator function. This produces MBD scores for both $\beta_0$ and $\beta_1$ dimensions.\\
    
    These topological features are combined with graph-theoretic features (Z-scores of node count, edge count, Randić index, and geometric-arithmetic index) to form the complete feature vector for classification.

\subsection{Katz Similarity Index}
We also explored the use of a modified version of the Katz similarity index in the prediction of the outcomes of tennis matches. This index comes from the following intuitive assumptions:
\begin{enumerate}
    \item If player X has defeated player Y in the past, then player X is more likely to defeat player Y again.
    \item If player X has defeated player Y and player Y has defeated player Z, we reasonably assume player X is more likely to defeat player Z
\end{enumerate}
These are the fundamental ideas behind our use of the Katz similarity index as a predictive measure for tennis matches.\\
Suppose that we have a graph with adjacency matrix $A$, and $i$ and $j$ are two vertices of the graph. Then the Katz similarity index $S_{katz}(i,j)$ is defined by
\[S_{katz}(i,j)=\sum_{k=1}^\infty \beta^kA^k[i,j]\]
where $\beta$ serves as an an exponential dampener. Note that $A^k[i,j]$ counts the number of paths from vertex $i$ to vertex $j$. Then the Katz similarity index (which from now on, we shall refer to as simply the Katz index), measures the number of paths between two vertices. If $\beta<1$, then this index gives greater weight to shorter paths and less weight to longer ones.\\
In order to modify this index for use in predicting tennis matches, we make a few changes. First, in order to give direction to our predictions, we wish to modify this index for digraphs. We would also like to add weights to edges to indicate both the number of wins as well as how relevant those wins are. In particular, we would like for more recent victories to matter more. It turns out that simply by replacing the adjacency matrix used in the formula for the Katz index by the adjacency matrix of a weighted digraph, we obtain exactly what we desire---an index that compiles information about the directed paths from $i$ to $j$ by weighting shorter paths more and giving more weight to more recent matches. We denote this modified Katz index on weighted digraphs by  $\overrightarrow{S}(i,j)$. We can think of this index as the advantage that player $i$ has over player $j$. By comparing the advantages of two players, we obtain a Katz score
\[\text{KatzScore}(i,j)=\overrightarrow{S}(i,j)-\overrightarrow{S}(j,i).\]
Then if $\text{KatzScore}(i,j)>0$, player $i$ has the advantage, and if $\text{KatzScore}(i,j)<0$ then player $j$ has the advantage.

\subsubsection{Making the Graph}
In order to use the KatzScore, we first have to create a graph on which we can use the score. There are many possible ways to make such a graph, but we make ours as follows in order to simplify computations. Note that we use sets in our dataset rather than matches since it gives us more granular data
\begin{itemize}
    \item We first create a node for each player in the dataset.
    \item If player X has ever defeated player Y in a set, we add an edge from player X to player Y.
    \item If we have an edge from player X to player Y, then we set its weight to be \[\sum \frac{1}{1+e^{1.5(c-\text{year})}}\] where the sum is taken over all sets in which player X defeated player Y, $c$ is the current year, and year is the year that the set was played. \\
\end{itemize}
This completes the construction of our graph.

\subsubsection{Making the Model}
    The first step to making the model was to finalize our modified Katz index. In particular, we had to find a cutoff for the summation (that is, a maximum path length to consider) as well as a value for $\beta$. Through comparing the predictive power of the index on samples for numerous combinations of these parameters, we determined that a cutoff of 4 and $\beta=0.3$ work well for predicting the outcomes of matches.\\
    Once we had these parameters set, we were able to set about making our predictive model. Since we are seeking to predict the outcome of a match using a continuous value, the KatzScore, we used a logistic fit for our predictive model.\\ 
    In order to obtain the model we used 5-fold cross validation on a sample of size 10000 (with 9023 usable points) to determine the parameters of a logistic fit curve. This sample was drawn from matches played between 2000 and 2022 inclusive. We excluded data from 2023-present to use to test the model. From our cross validation, we obtained the following model
    \[\text{P(Player 1 wins)} = \frac{1}{1+e^{-(-0.004794+(0.429513\cdot\text{KatzScore}))}}\]
    where KatzScore is the KatzScore between Player 1 and Player 2. This model returns the probability that Player 1 wins. Note that if the KatzScore is 0, the model returns a probability of $0.4988$ which is very close to the desired value of $0.5$. 


\section{Experimental results}

We evaluated our methods on ATP singles match data as described at the beginning of Section \ref{methods}. This section presents results for both the lower-star filtration approach and the Katz similarity index method.

\subsection{Lower-Star Filtration}
We begin by describing our initial implementation with limited data, followed by the optimizations that enabled large-scale analysis and comprehensive results.

\subsubsection{Initial Implementation} \label{initial}

In our initial implementation, we faced computational constraints that limited our analysis to a small subset of the available data. We randomly sampled approximately 2\% of matches from the complete dataset, yielding 1,200 matches for model development. This subset was divided into training (70\%, $n=840$) and testing (30\%, $n=360$) sets. We also evaluated the model on an external validation set from the 2025 Vanda Pharmaceutical Hellenic Championship tournament.\\

We trained three classification algorithms on the topological and graph-theoretic features extracted from the induced subgraphs. The performance of each model is summarized in Table~\ref{tab:initial_models}.

\begin{table}[H]
	\centering
	\caption{Performance comparison of classification models on initial dataset (1,200 matches).}
	\label{tab:initial_models}
	\begin{tabular}{lcc}
		\hline
		\textbf{Model} & \textbf{Accuracy} & \textbf{Threshold ($\alpha$)} \\
		\hline
		Logistic Regression & 58.82\% & 0.27 \\
		Random Forest & 60.78\% & 0.27 \\
		XGBoost & \textbf{61.27\%} & 0.27 \\
		\hline
	\end{tabular}
\end{table}

XGBoost emerged as the best-performing model with an accuracy of 61.27\%. Notably, using cross-validation, optimal performance was achieved at a decision threshold of $\alpha = 0.27$ rather than the standard $\alpha = 0.5$, where any predicted probability $p > 0.27$ indicates that the player with the higher $g(v)$ value is predicted to win. This lower threshold addresses the class imbalance introduced by our directional encoding based on $g(v)$ values. Compared to an XGBoost model using the default threshold of $\alpha = 0.5$ (which achieved 55\% accuracy), the optimized threshold provided an improvement of approximately 6.27 percentage points.\\

To assess generalization beyond our training data, we applied the trained XGBoost model to predict all matches from the 2025 Vanda Pharmaceutical Hellenic Championship. Of the 39 total matches in the tournament, 35 could be predicted (4 matches were excluded because one or both players lacked sufficient historical data in our dataset). The model achieved an accuracy of 57.14\% (20 correct predictions out of 35) on this external validation set.\\

While this accuracy is above random chance (50\%), the performance decrease from 61.27\% (test set) to 57.14\% (external tournament) suggests some overfitting or dataset-specific patterns. Additionally, the small sample size (35 matches) provides limited statistical power for robust evaluation. These initial results highlighted the need for larger training datasets to improve generalization and statistical reliability, as well as computational optimizations to enable analysis of the complete match dataset. The following subsection describes the algorithmic optimizations that enabled us to overcome these constraints and analyze substantially larger datasets.

\subsubsection{Optimization Strategy, Feature Analysis and Model Comparison}

\paragraph{Computational Challenges and Optimizations}

The initial implementation described in Section~\ref{initial} faced severe computational bottlenecks. The $g$-function computation used numpy array searches with \texttt{np.where()}, resulting in $O(n^2)$ complexity for player lookups across the dataset. Finding the competitive neighborhood around two players relied on enumerating all simple paths using \texttt{nx.all\_simple\_paths()}, which has exponential complexity $O(\exp(\text{cutoff}))$. Additionally, each induced subgraph required triangle enumeration for 2-simplices, adding $O(k^3)$ complexity where $k$ is the number of nodes in the subgraph. At this rate, processing the complete dataset of 65,834 matches would require prohibitively long computation time on our available hardware.\\

To enable large-scale analysis, we implemented three key optimizations that collectively reduced per-match computation time. First, we replaced numpy array operations with Python dictionaries for player statistics accumulation, reducing player lookup complexity from $O(n)$ to $O(1)$. Second, we replaced the exponential path enumeration algorithm with an ego graph intersection method. For two players $p_1$ and $p_2$, we extract the $r$-neighborhood (ego graph) around each player, compute the union of these neighborhoods, and create the induced subgraph from the combined vertex set
\[
	H = G[V(\text{ego}(p_1, r)) \cup V(\text{ego}(p_2, r))]
\]
This reduces complexity from $O(\exp(\text{cutoff}))$ to $O(|V|^r)$. Third, we removed 2-simplex (triangle) computation from the filtration process, computing only $\beta_0$ (connected components) and $\beta_1$ (loops) instead of all Betti numbers. For tennis prediction, $\beta_0$ and $\beta_1$ capture the most relevant topological features while avoiding the computational expense of triangle enumeration.\\

Finally, we implemented a two-tier caching system to avoid redundant computation. The first tier stores player-level persistence diagrams (\texttt{player\_cache\_all.pkl}), while the second tier stores centrality measures (\texttt{centralities\_cache.pkl}). Since many players appear in multiple matches, caching their ego graph computations and centrality values provides substantial time savings throughout the processing pipeline. \\

The optimization computations were performed on a laptop equipped with an Intel Core Ultra 9 285H processor (2.90 GHz) and 64 GB RAM. The total execution time was approximately 242 hours, with an average processing time of 13.18 seconds per match.

\paragraph{Large-Scale Dataset and Feature Engineering}

With optimized algorithms, we processed a larger dataset of 65,834 matches for comprehensive analysis. For each match between players $p_1$ and $p_2$, we computed 46 features across six categories. The \textit{base graph features} include the number of nodes and edges in the induced subgraph, along with two topological indices: the Randi\'c index $\sum_{(u,v) \in E} \frac{1}{\sqrt{\deg(u) \cdot \deg(v)}}$ and the geometric-arithmetic index $\sum_{(u,v) \in E} \frac{2\sqrt{\deg(u) \cdot \deg(v)}}{\deg(u) + \deg(v)}$. The \textit{$g$-function difference} is defined as $\Delta g = g(p_1) - g(p_2)$, where $g(v)$ is the normalized player value from Section~\ref{gfunction}. The \textit{ranking difference} is $\Delta \text{rank} = \text{rank}(p_1) - \text{rank}(p_2)$. The \textit{centrality differences} include degree, closeness, betweenness, and PageRank differences between the two players.\\

For TDA features, we computed two types of representations for each of four summary methods (VAB, HNAV, HWNAV, OW-HNPV) across two homological dimensions ($\beta_0$, $\beta_1$). The \textit{Modified Band Depth (MBD)} features measure the statistical depth of each match's topological signature relative to a reference distribution. The \textit{direct statistics} features include the mean, maximum, and standard deviation of each topological summary function. During feature analysis, we discovered that all 16 features related to $\beta_1$ (1-dimensional homology representing loops/cycles) exhibited zero variance across the entire dataset. This occurred because the triangle elimination in our optimization step prevents the formation of persistent cycles in the induced subgraphs. Consequently, these $\beta_1$-related features were excluded from modeling, resulting in a final feature set of 26 variables.

\begin{center}
\begin{figure}[H] 
    \centering
    \includegraphics[scale = .4]{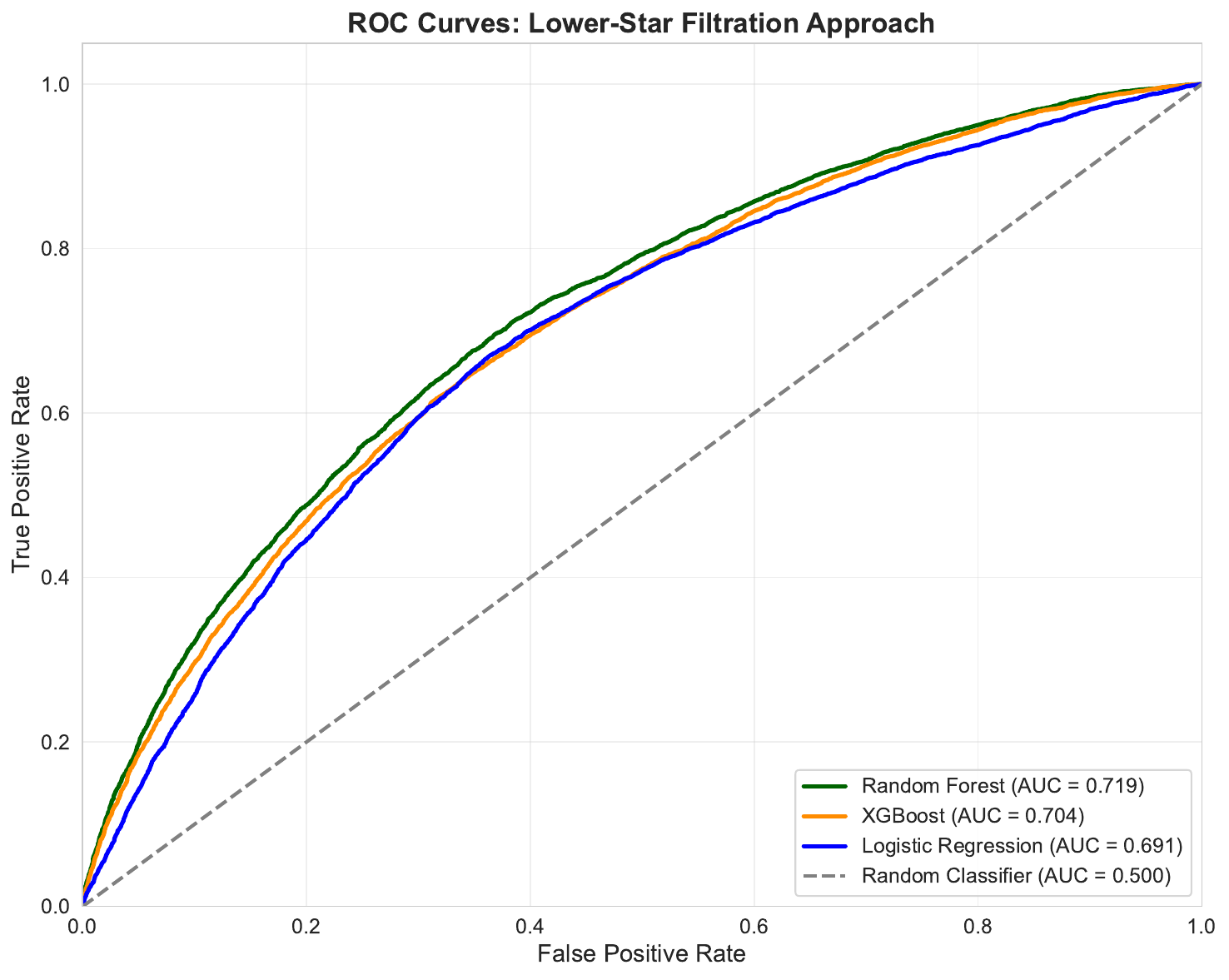}
    \caption{ROC curves comparing Random Forest, XGBoost, and Logistic Regression models for the lower-star filtration approach. The Random Forest model achieves the strongest overall discriminative performance.}
    \label{model_comparison}
\end{figure}
\end{center}

\paragraph{Model Performance and Comparison}
We evaluated three classification algorithms on the 26-feature set using a 70/30 train-test split with stratified sampling to preserve class balance. The results are shown in Table~\ref{tab:model_comparison} and Figure \ref{model_comparison}.

\begin{table}[H]
	\centering
	\caption{Performance comparison of classification models on large-scale dataset (65,885 matches).}
	\label{tab:model_comparison}
	\begin{tabular}{lccccc}
		\hline
		\textbf{Model} & \textbf{Acc.} & \textbf{AUC} & \textbf{Prec.} & \textbf{Rec.} & \textbf{F1} \\
		\hline
		Logistic Regression & 0.652 & 0.691 & 0.652 & 0.652 & 0.652 \\
		Random Forest & \textbf{0.662} & \textbf{0.719} & \textbf{0.664} & \textbf{0.657} & \textbf{0.660} \\
		XGBoost & 0.661 & 0.718 & 0.665 & 0.650 & 0.658 \\
		\hline
	\end{tabular}
\end{table}

The Random Forest model (with 500 estimators, maximum depth of 10, and random state 42) achieved the best overall performance with 66.2\% accuracy and AUC of 0.719. This represents a substantial improvement of 4.9 percentage points over our initial implementation on a smaller dataset of 1,200 matches (61.3\% accuracy). The improved performance is attributed to both the significantly larger training dataset (65,885 unique matches) and optimized feature engineering pipeline that eliminated duplicate entries and zero-variance features.

\begin{center}
\begin{figure}[H] 
    \centering
    \includegraphics[scale = .4]{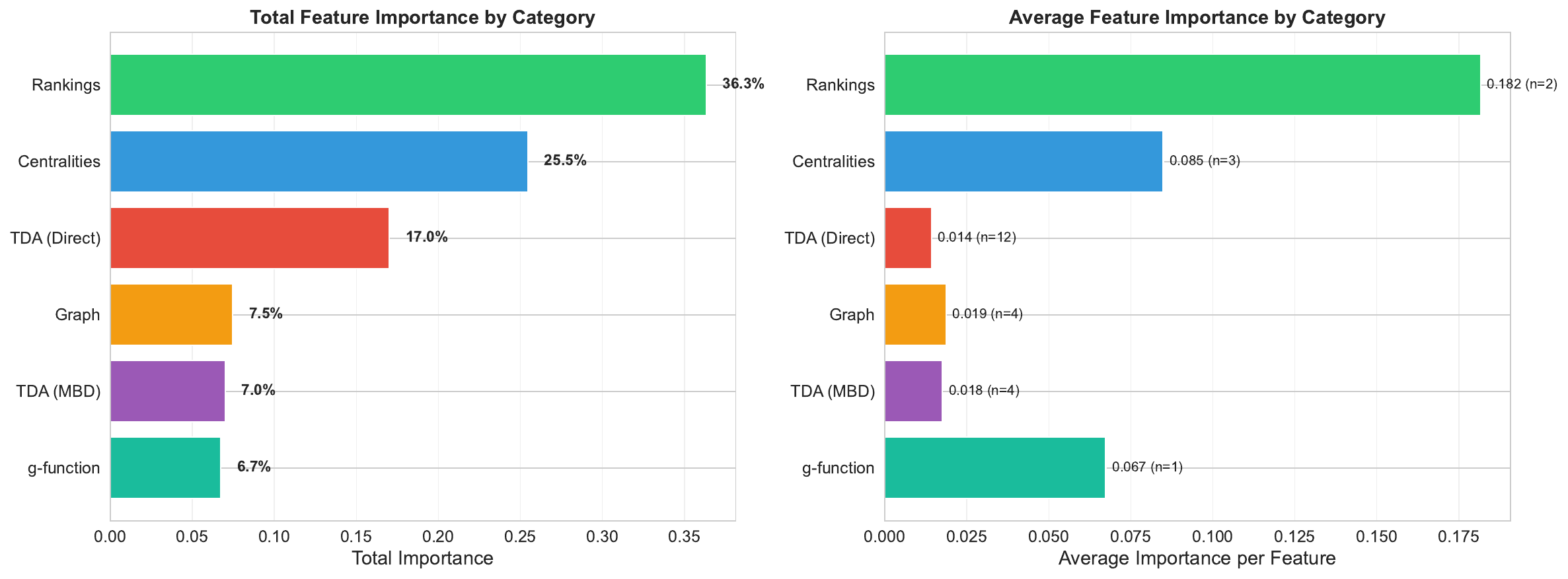}
    \caption{Aggregate and average feature importance by category for the Random Forest model. Topological (TDA) features contribute a substantial portion of total importance but exhibit overlap with traditional graph-based and ranking features.}
    \label{fig:importance_category}
\end{figure}
\end{center}

\paragraph{Feature Importance Analysis}
Feature importance analysis from the Random Forest model revealed the relative contribution of different feature categories to the model's decision-making process. As shown in Figure \ref{fig:importance_category}, the ranking features (rank difference and g-function difference) collectively account for 36.3\% of predictive power, confirming that official ATP rankings effectively capture player strength. Graph centrality features collectively contribute 25.5\% of importance, indicating that a player's position within the competitive network strongly influences match outcomes. Combined TDA features (MBD + direct statistics) contribute 24.0\% total importance, with direct statistics (17.0\% total) substantially outperforming MBD depth scores (7.0\%), indicating that simple summary statistics (mean, max, std) of topological functions are more predictive than depth-based representations for this application. Among the TDA methods, VAB contributed the highest importance (7.3\% total), followed by HWNAV (5.9\%), HNAV (6.0\%), and OW-HNPV (4.9\%).

To assess dependence on external data, we evaluated model performance excluding the ranking feature. The topology-only configuration achieves 63.56\% accuracy (AUC = 0.691), a 2.52 percentage point decrease from the full model. When rankings are removed, centrality importance increases from 25.5\% to 45.5\% and $g$-function contribution rises from 6.7\% to 13.4\%, indicating these features partially substitute for ranking information. This demonstrates that topological and graph-theoretic features alone achieve above-chance prediction without external data, while rankings provide complementary signal that improves performance when available.

\paragraph{TDA Method Comparison}
To systematically evaluate different TDA summarization approaches, we conducted two controlled experiments. First, we compared MBD depth scores with direct statistics of TDA summaries. The results in Table~\ref{tab:mbd_vs_direct} show that direct statistics slightly outperform MBD depth scores, while the combined approach offers no additional benefit, suggesting redundancy between the two representations.

\begin{table}[H]
	\centering
	\caption{Comparison of TDA representation methods.}
	\label{tab:mbd_vs_direct}
	\begin{tabular}{lccc}
		\hline
		\textbf{Configuration} & \textbf{\# Features} & \textbf{Accuracy} & \textbf{AUC} \\
		\hline
		Baseline (no TDA) & 10 & 0.660 & 0.718 \\
		MBD Depth Only & 14 & 0.662 & 0.718 \\
		Direct Statistics Only & 22 & \textbf{0.662} & \textbf{0.719} \\
		MBD + Direct (Combined) & 26 & 0.662 & 0.719 \\
		\hline
	\end{tabular}
\end{table}

\begin{center}
\begin{figure}[H] 
    \centering
    \includegraphics[scale = .4]{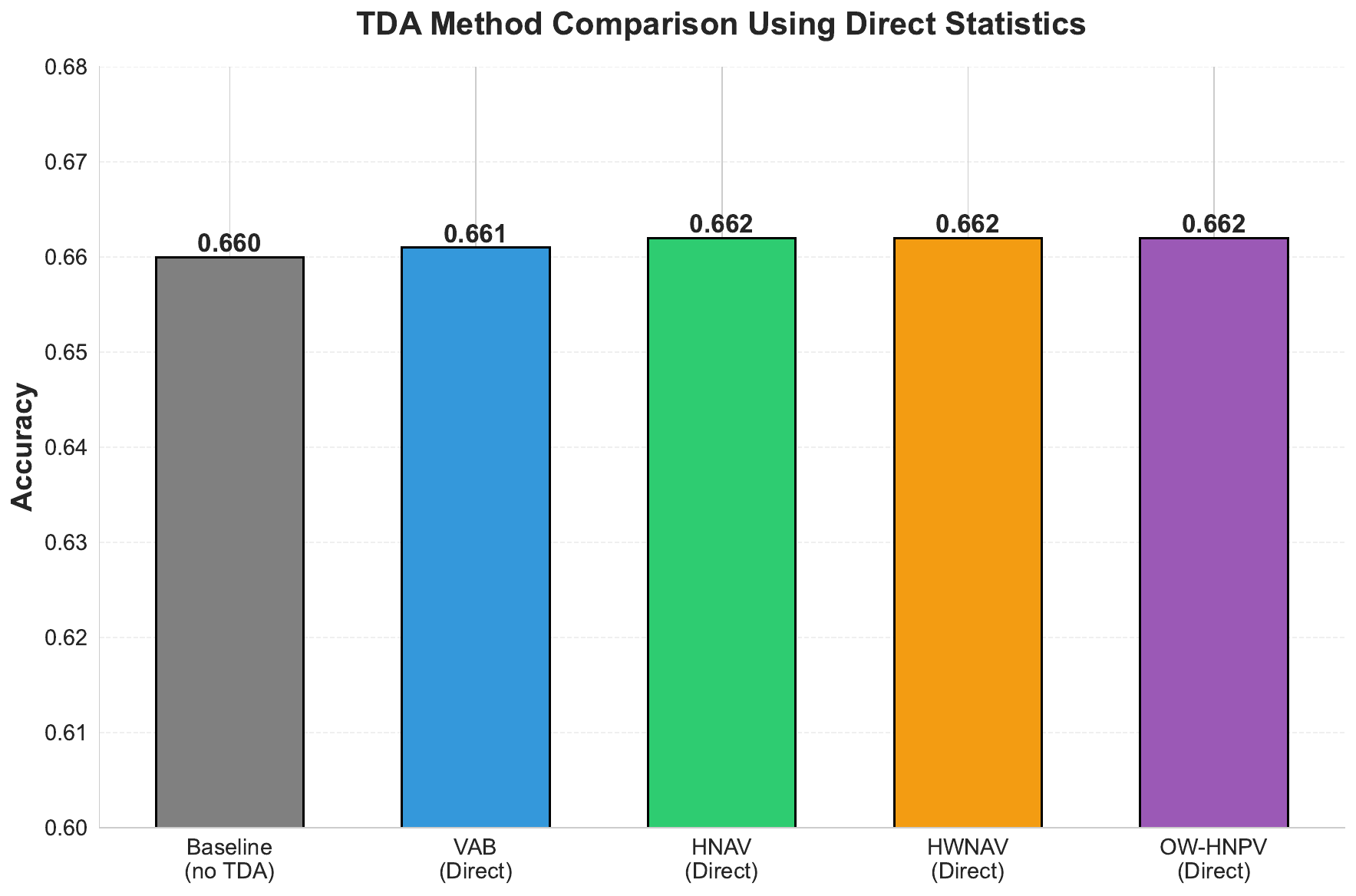}
    \caption{Comparison of predictive accuracy for different TDA summary methods using direct statistics. All TDA-augmented models match or slightly outperform the baseline model, with the OW-HNPV summary achieving the highest accuracy.}
    \label{fig:tda_methods}
\end{figure}
\end{center}

Second, we compared four TDA summary methods,VAB (Vector of Averaged Bettis), HNAV (Hierarchical Normalized Averaged Velocity), HWNAV (Hierarchical Weighted Normalized Averaged Velocity), and OW-HNPV (Our Weighted Hierarchical Normalized Persistence Velocity), using direct statistics representations. Figure~\ref{fig:tda_methods} shows that all four TDA methods achieved comparable performance, with OW-HNPV, HWNAV, and HNAV tied at the highest standalone accuracy (66.18\%), though VAB contributed the most feature importance in the combined model (7.3\%).

\subsection{Katz Index}

We now present results for our graph-theoretic approach based on the Katz similarity index.

\subsubsection{Training Results}
We trained our model using 5-fold cross-validation on the data from 2000 to 2022 leaving out the data from 2023 to 2025 for testing. On the training data, we had the following results.


\begin{table}[H]
	\centering
	\caption{Summary statistics of model performance across all evaluated matches.}
	\label{tab:model_summary}
	\begin{tabularx}{0.85\textwidth}{
		>{\centering\arraybackslash}X
		@{\hspace{1.5cm}}
		>{\centering\arraybackslash}X
	}
		\hline
		\textbf{Metric} & \textbf{Value} \\
		\hline
		Sample Size & 9{,}023 \\
		Mean Accuracy & 0.6461 \\
		Accuracy Standard Deviation & 0.0131 \\
		Mean AUC & 0.6911 \\
		AUC Standard Deviation & 0.0159 \\
		Mean Log Loss & 0.6494 \\
		\hline
	\end{tabularx}
\end{table}

\subsubsection{Testing Results}
We tested our model on all sets played between 2023 and 2025 inclusive. We break down our results by year and also provide overall results. 


\begin{table}[H]
	\centering
	\caption{Year-wise model performance comparison.}
	\label{tab:yearly_performance}
	\begin{tabularx}{1\textwidth}{
		>{\centering\arraybackslash}X
		@{\hspace{0.5cm}}
		>{\centering\arraybackslash}X
		@{\hspace{0.5cm}}
		>{\centering\arraybackslash}X
		@{\hspace{0.5cm}}
		>{\centering\arraybackslash}X
		@{\hspace{0.5cm}}
		>{\centering\arraybackslash}X
	}
		\hline
		\textbf{Metric} & \textbf{2023} & \textbf{2024} & \textbf{2025} & \textbf{Overall} \\
		\hline
		Population Size & 6{,}863 & 6{,}911 & 5{,}488 & 19{,}262 \\
		Accuracy & 0.6193 & 0.6324 & 0.6221 & 0.6248 \\
		AUC & 0.6601 & 0.6767 & 0.6604 & 0.6659 \\
		Log Loss & 0.6553 & 0.6538 & 0.6561 & 0.6550 \\
		\hline
	\end{tabularx}
\end{table}

We observe that the model performs pretty consistently over the years with very similar results in 2023, 2024, and 2025. And these results are very close to the results from the training data as well. This suggests that the model has consistent predictive capability and is able to consistently predict the outcomes of unknown matches better than random chance.\\

Now, the reader might have considered that between two players there may not exist a path in the digraph that we had created for our Katz Index-based approach. This is indeed true. When this is the case, the KatzScore of those two players is zero. In order to examine whether the model fared any better when we excluded such sets in our testing, we excluded these sets and tested the model again. There is virtually no change. We omit the testing results for this because all metrics are within one ten-thousandth of the metrics listed in the table above.

\section{Conclusion}

We presented two novel approaches for predicting tennis match outcomes using topological data analysis and graph-theoretic methods. Our lower-star filtration approach with Random Forest achieved 66.2\% accuracy, while the modified Katz similarity index achieved 62.48\%, both substantially outperforming random prediction. This work makes several contributions: (1) first application of lower-star filtration to tennis prediction, (2) systematic comparison of four TDA summary methods in sports analytics, and (3) algorithmic optimizations enabling large-scale TDA computation.\\

Our results provide an important methodological insight. While TDA successfully extracts topological features from competitive networks (contributing 24\% model importance), these features are largely redundant with simpler graph-theoretic measures and official rankings. This suggests that for tennis prediction specifically, the sophisticated ATP ranking system and basic centrality measures already capture most network structure. The modest performance gain (0.2\% accuracy improvement over baseline) indicates that TDA's value may lie more in interpretability and theoretical understanding than predictive improvement for this application. This finding has broader implications for sports analytics: when domain-specific ranking systems are highly developed (as in tennis with ATP rankings), topological methods may offer diminishing returns for pure prediction tasks, though they remain valuable for understanding competitive structure.

\paragraph{Limitations}

The lower-star filtration approach has several inherent limitations. First, it cannot generate predictions for players with insufficient historical match data. New players entering professional competition or players returning from extended absences cannot be analyzed until they accumulate sufficient match history in our dataset. Second, the temporal weighting scheme may systematically bias predictions toward younger players by emphasizing peak performance years, potentially undervaluing long-term consistency in veteran players. Third, all loop-based topological features ($\beta_1$) exhibited zero variance under our optimized construction due to the absence of triangles in the filtration, leaving unresolved whether higher-order cyclical structures provide predictive value at larger scales or in global network analysis.\\

The model's reliance on ATP rankings, while providing meaningful predictive value (36.3\% combined importance for ranking-based features), introduces operational complexity for real-world deployment. ATP rankings are updated weekly, requiring either model retraining or feature recalculation before each prediction to maintain accuracy. In production environments, this necessitates automated data pipelines to fetch current rankings, update features, and regenerate predictions on a regular schedule. While such automation is feasible using ATP's publicly available ranking data, it introduces dependencies on external data sources and update schedules. Our analysis demonstrates that the model maintains reasonable performance (63.56\% accuracy) using only network-derived features when rankings are unavailable, providing a degraded but functional fallback mode. However, the 2.52 percentage point accuracy improvement from rankings justifies the additional implementation complexity for applications where maximum predictive accuracy is required.\\

The Katz similarity approach faces complementary limitations. Data sparsity in the competitive network poses challenges, as disconnected player pairs yield zero similarity scores that provide no informative signal. This issue is particularly pronounced in smaller tournaments and early-career matches where players have fewer historical interactions. Additionally, while computing match-specific Katz scores by removing future matches from the graph would provide greater temporal precision, this is computationally expensive; our use of historical snapshots (including only matches before the prediction date) mitigates this cost but sacrifices some temporal granularity. Finally, both methods capture only network structure and player strength, omitting contextual factors such as playing surface, tournament importance, weather conditions, recent form, and psychological dynamics that may influence match outcomes.

\paragraph{Future Directions}

Future work should address these limitations through several avenues. For players with limited match history, transfer learning or similarity-based approaches could leverage information from comparable players based on physical attributes, playing style, or junior tournament records. Alternative temporal weighting schemes that account for opponent strength and career trajectories could reduce bias against veteran players. Incorporating surface-specific features and tournament context could improve prediction accuracy, as could explicit modeling of head-to-head records and recent form. Analyzing the complete 66,000 match dataset with triangle computation enabled would resolve whether $\beta_1$ and $\beta_2$ features provide additional predictive value beyond connectivity structures. Finally, deep learning approaches, particularly graph neural networks, could potentially capture complex nonlinear interactions between topological, graph-theoretic, and contextual features, though careful optimization would be necessary to balance model complexity against computational efficiency and interpretability.

 \section*{Acknowledgments}
    This research was conducted as part of Topics in Graph Machine Learning (STAT 391) at the University of Evansville during Fall 2025. The work represents a collaborative effort between the instructor and students in the course. We thank Lily Dwyer for her comments during class discussions. 
    Claude (Anthropic) was used to assist with debugging Python code during the optimization phase of the lower-star filtration implementation and to polish portions of the manuscript. All research design, methodological development, experimental work, data analysis, and intellectual content are the original contributions of the course participants.

\end{document}